\documentclass[twoside, twocolumn]{article}

\usepackage[english]{babel}

\usepackage[sc]{mathpazo}

\usepackage[T1]{fontenc}
\usepackage{microtype}
\usepackage[margin=2cm, hmarginratio=1:1, top=32mm, columnsep=20pt]{geometry}
\usepackage[hang, small, labelfont=bf, up, textfont=it, up]{caption}
\usepackage{booktabs}
\usepackage{lettrine}

\usepackage{amsmath}

\usepackage{enumitem}
\setlist[itemize]{noitemsep}

\usepackage{abstract}

\usepackage{titlesec}
\renewcommand\thesection{\Roman{section}}
\renewcommand\thesubsection{\roman{subsection}}
\titleformat{\section}[block]{\large\scshape\centering}{\thesection.}{1em}{}
\titleformat{\subsection}[block]{\large}{\thesubsection.}{1em}{}

\usepackage{fancyhdr}
\usepackage{titling}
\usepackage{hyperref}
\usepackage{subcaption}
\usepackage{dblfloatfix}
\usepackage{floatrow}
\usepackage{multirow}
\usepackage{graphicx}
\graphicspath{ {images/} }

\pretitle{\begin{center}\Huge\bfseries}
\posttitle{\end{center}}
\title{LIGAR: Lightweight General-purpose Action Recognition}
\author{
\textsc{Evgeny Izutov} \\[1ex]
\normalsize Intel Corporation \\
\normalsize \href{mailto:evgeny.izutov@intel.com}{evgeny.izutov@intel.com}
}
\date{}

\renewcommand{\maketitlehookd}{
\begin{abstract}
\noindent Growing amount of different practical tasks in a video understanding problem has addressed the great challenge
aiming to design an universal solution, which should be available for broad masses and suitable for the demanding
edge-oriented inference. In this paper we are focused on designing a network architecture and a training pipeline to
tackle the mentioned challenges. Our architecture takes the best from the previous ones and brings the ability to be
successful not only in appearance-based action recognition tasks but in motion-based problems too. Furthermore, the
induced label noise problem is formulated and Adaptive Clip Selection (ACS) framework is proposed to deal with it.
Together it makes the LIGAR framework the general-purpose action recognition solution. We also have reported the
extensive analysis on the general and gesture datasets to show the excellent trade-off between the performance and the
accuracy in comparison to the state-of-the-art solutions. Training code is available at:
\url{https://github.com/openvinotoolkit/training_extensions}. For the efficient edge-oriented inference all trained
models can be exported into the OpenVINO\texttrademark format.
\end{abstract}
}

\begin{document}

\maketitle


\section{Introduction}

\lettrine[nindent=0em,lines=3]{N}{}owadays Action Recognition (AR) plays the key role in many real-world applications,
including human-robot interaction, behavior analysis, gesture translation. It aims to solve the classification problem
in the video domain. The recent success of Deep Learning (DL) based solutions has brought the ability to solve the
mentioned above problems close to the human level. Unfortunately, there is a big gap between a working solution developed
for the academic purposes and an edge-oriented network design.

To connect the business goals and academic achievements the researchers have proposed several solutions for adverted
tasks, like MobileNets~\cite{MN-V1, MN-V2, MN-V3} for 2D image classification and the OSNet architecture~\cite{OSNet}
for the Re-Identification problem. The AR domain still lacks the fast and edge-oriented solutions. So, we developed the
lightweight action recognition network architecture called \textbf{LIGAR} (\textbf{LI}ghtweigt  
\textbf{G}eneral-purpose \textbf{A}ction \textbf{R}ecognition) to fill the gap between accurate and real-time solutions.

\textbf{Motivation.} Current SOTA Action Recognition approaches mostly use 3D-like backbone design~\cite{I3D, P3D,
R(2+1)D, S3D, X3D} to capture the motion component paired with appearance one. It allows us to ingest huge datasets
(e.g. Kinetics~\cite{Kinetics} and YouTube-8M~\cite{Ytb8m}) and demonstrates the impressive accuracy on the admissible
robustness level. Unluckily the final 3D-like AR networks are too big for the edge-oriented inference and cannot
be used somewhere except benchmarks.

The opposite segment of fast AR networks is represented by 2D network architectures~\cite{TSN, TSM, TRN}, which split
a video processing pipeline on independent image processing and the late feature fusion. The 2D networks work fast
enough for the real-time inference. The main drawback of the adverted 2D solutions is an inability to capture complex
motions, like hand movements in the sign language recognition problem (e.g. American Sign Language -- ASL).

Summarizing, the development of a general-purpose network requires capturing the motion component as well as appearance
one. In the proposed paper we are focused on merging representatives of 2D and 3D worlds of lightweight architectures --
MobileNet-V3~\cite{MN-V3} and X3D~\cite{X3D} respectively. The obtained architecture is suitable for motion capturing
mostly and to enhance processing the appearance component the model is supplemented with additional global branch
(proposed in the LGD paper~\cite{LGD}).

\textbf{Our contributions} are as follows:
\begin{itemize}
\item Extending the family of efficient 3D networks for the general-purpose action recognition by merging X3D~\cite{X3D}
and LGD~\cite{LGD} video processing frameworks with the lightweight edge-oriented MobileNet-V3~\cite{MN-V3} backbone
architecture.
\item Supplementing the training procedure with a bag of tricks (augmentations, multi-head pre-training, feature
regularization, metric-learning based head, learning rate scheduler) to deal with limited size of AR datasets to reach
robustness.
\item Proposing the Adaptive Clip Selection (ACS) module to tackle with a weak temporal annotation in common AR
datasets.
\end{itemize}

In addition, we release the training framework\footnote{\url{https://github.com/openvinotoolkit/training_extensions}}
that can be used in order to re-train or fine-tune our model with a custom database. The final model is inference-ready
to use with Intel$^{\textregistered}$ OpenVINO\texttrademark
toolkit\footnote{\url{https://software.intel.com/en-us/openvino-toolkit}} -- the sample code on how to run the model in
the demo mode is available at Intel$^{\textregistered}$ OpenVINO\texttrademark
OMZ\footnote{\url{https://github.com/openvinotoolkit/open_model_zoo}}.

\textbf{Index terms:} action recognition, network regularization, lightweight network, edge-oriented inference, metric
learning head, label noise suppression.


\section{Related Work}

\subsection{Action Recognition}

\lettrine[nindent=0em,lines=3]{I}{}n the Action Recognition problem we are concentrated on solving the classification
problem on a sequence of images, which can be derived from a live video stream or an untrimmed long video. First
attempts to deal with it were based on extending the 2D classification networks into the 3D use case by addition an extra
temporal dimension inside the CNN's kernels, like in C3D~\cite{C3D} and I3D~\cite{I3D} papers. The final networks were
too large for the existed datasets and suffered from strong over-fitting. In the same time, purely 2D solutions perform
independent image processing and after this the extracted features are merged at the decision~\cite{TwoStream, TSN} or
feature~\cite{TRN} levels. That design allows the 2D network to be the fastest video processing framework but not
accurate.

The latter improvement is addressed to the publication of sizable datasets, like Kinetics~\cite{Kinetics} and
YouTube-8M~\cite{Ytb8m}, which are suitable for pre-training purposes. Next steps were focused on the reduction of
parameters of 3D-based networks (e.g. R(2+1)D~\cite{R(2+1)D}, S3D~\cite{S3D}), extending 2D approaches with Graph
Convolutional Networks (GCNs)~\cite{TRG} and merging 2D and 3D networks into the single one (e.g. SlowFast
\cite{SlowFast}, AssembleNet~\cite{AssembleNet}, LGD~\cite{LGD}) to achieve the trade-off between the accuracy of 3D
networks and the low computation budget of 2D ones. We pursue the the same paradigm and designed the LGD-based
architecture.

Other researchers pay attention to integrating specialized attention modules into the backbone~\cite{RCCA} or feature
aggregation module~\cite{NL} or designing a specialized head for better spatio-temporal feature aggregation~\cite{Bert}.
Unfortunately, the listed above methods lead to the grow of computations and thereby are less suitable for the
edge-oriented inference.

\subsection{Edge-oriented inference}

\lettrine[nindent=0em,lines=3]{A}{}fter the resounding success of DL-based approaches of solving a wide range of
complex problems the necessity of porting solutions closer to the end users by adopting networks for the inference on
the edge has appeared. The first 2D solutions are represented by the MobileNet family~\cite{MN-V1, MN-V2, MN-V3}, the
ShuffleNet architecture~\cite{ShuffleNet-V1, ShuffleNet-V2} and EfficientNets~\cite{EfficientNet} later.

The attempts to bring 3D networks closer to business include mostly introducing the X3D~\cite{X3D} and MoViNet
\cite{MoViNet} network families. The last one is designed automatically by the NAS-based approach. In the proposed paper
we adhere to the X3D framework and merge it with the mentioned above edge-oriented 2D backbone -- MobileNet-V3
architecture.

\begin{figure*}[t]
\centering
\includegraphics[width=\textwidth]{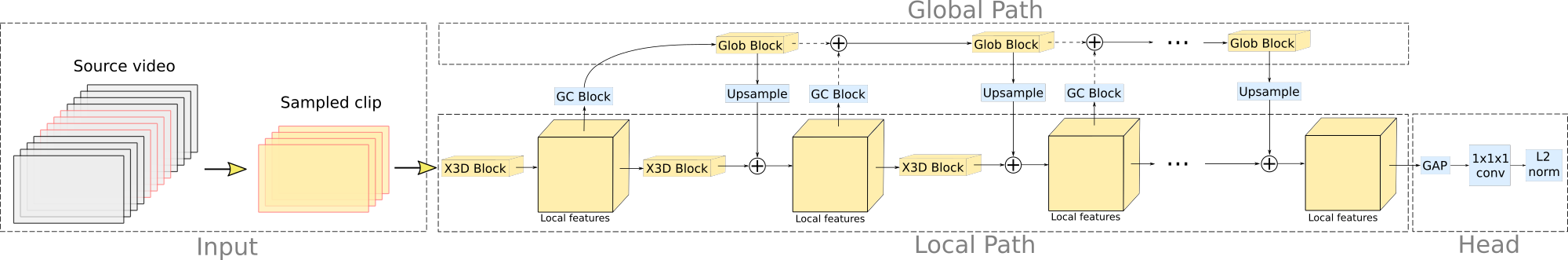}
\caption{Overall network design. The model consists of Local and Global paths with intermediate fusion modules like in
LGD network \cite{LGD}. The Local path is built upon X3D MobileNet-V3-Large backbone architecture. The Global path
follows the original LGD design but with Global Context (GC) blocks instead of Global Average Pooling (GAP) operator.
The lightweight head is represented by GAP operator and $1 \times 1 \times 1$ convolution module. The network output is
$l_2$ normalized to enable metric-learning losses.}
\label{fig:architecture}
\end{figure*}

\subsection{Network regularization}

\lettrine[nindent=0em,lines=3]{A}{}nother question addressed to the researchers is the ability to train a network on the
limited amount of a data without significant over-fitting. The most common way to tackle it is to use advanced
augmentation techniques, like MixUp~\cite{MixUp}, AugMix~\cite{AugMix}, CrossNorm~\cite{CrossNorm} and so on.

A different point of view is related to using feature-level regularization methods, like Dropout family
\cite{Dropout, ContinuousDropout, InfoDrop, FocusedDropout}, mutual learning~\cite{MutualLearning} and self-feature
regularization~\cite{SFR}. We follow the mutual learning paradigm and blend it with the mentioned above AugMix strategy.

Further, the network regularization can be performed by using more sophisticated losses instead of default one (the pair
of the Softmax normalization and the CrossEntropy loss): transition to the Metric-Learning (ML) paradigm~\cite{ASLNet}
and using AM-Softmax loss~\cite{AM-Softmax} as a main target and some auxiliary losses from the Re-Identification task,
like PushPlus~\cite{RMNet} and CenterPush~\cite{ASLNet} losses. We act in accordance with the same practices and use ML
head with the aforementioned auxiliary losses.


\section{Network design}

\subsection{Motion-appearance trade-of}

\lettrine[nindent=0em,lines=3]{O}{}riginally the data sources for AR can be divided by the data representation onto
continual and sparse. The last one is most common for general-purpose scenarios and described by appearance mostly
rather than motion. In other words, the significant amount of data samples from that category can be viewed as a
sequence of key frames, where each frame reflects a change in a global appearance. For better understanding imagine the
"covering something" atomic action -- to recognize the mentioned class it is needed to extract the sequence of two
frames only -- "uncovered object" and then "covered by something" one. As it can be seen the motion component in the
above sample is rudiment. To recognize similar classes it is enough to model the relationship between key frames by a 2D
network, like in TSN~\cite{TSN} or TRN~\cite{TRN}.

The continues segment of data representations consists of gesture recognition scenarios, like ASL (American Sign
Language) recognition task. It is complicated by the difficult hand and fingers movements. The attempt to compress an
image sequence into a sequence of key frames (comparable with 2D approaches) fails~\cite{MSASL} due to the unchanging
appearance component (a person or background on a video) from the one hand and the rapidly changing motion component 
(fingers) on the other hand. The only way to model the motion component is using fully 3D networks.

In this paper we accept the common paradigm for "inflating" 2D network architectures into the 3D one, while keeping the
original 2D weight initialization, as it was proposed in I3D~\cite{I3D}. In terms of edge-oriented network design there
are two main inflating frameworks: S3D~\cite{S3D} and recently published X3D~\cite{X3D}. The choice between S3D and X3D
frameworks can be viewed as the choice between the motion- and the appearance-preferable architectures respectively due
to the different placement of a temporal convolution kernel -- inside a depth-wise kernel (X3D) and after it (S3D). As
it was shown in~\cite{ASLNet} the S3D framework is most suitable for the heavy gesture recognition scenarios, like the
ASL gesture recognition problem. But our extensive experiments evinced that S3D is worse than X3D framework in general
scenarios (e.g. UCF-101~\cite{UCF101} or AcivityNet~\cite{ActivityNet} datasets). Eventually, we have concluded to use
the X3D framework.

\subsection{Multi-path architecture}

\lettrine[nindent=0em,lines=3]{R}{}ecently proposed Local and Global Diffusion (LGD) framework~\cite{LGD} achieves SOTA
results on several general-purpose datasets. The main idea of the mentioned method is to split the regular single-path
backbone design into two paths for separately processing the local and global context. The framework also offered the
between paths communication modules. We found the proposed framework is preferable in terms of motion-appearance trade-off
due to the possibility to move the lion's share of a work of appearance component processing into the lightweight global
path. Such a find also allows us to compensate the loss for the gesture recognition issues, while making the choice in
favor X3D instead of S3D framework.

In our paper we implemented the same LGD-based backbone architecture with several changes:
\begin{itemize}
\item We refused of using the kernel function to merge the global and local branches. Instead of it the local branch is
used as the backbone output and the global branch is shorter by single LGD block. We did not beneficiate from using any
complex merging scheme because of following the end-to-end training paradigm instead of separate path pre-training,
proposed originally (see section~\ref{pre-training}).

\item Basically, the local-to-global diffusion module uses Global Average Pooling (GAP) operator to merge
spatio-temporal representations into the single feature-vector. Experimentally we have found that the more accurate way
is to use the attention module through the spatio-temporal dimensions to extract the most relevant features. Having
multiple choice of possible architectures of the attention module we have focused on Global Context (GC) block
\cite{GCN}.
\end{itemize}

\subsection{Overall architecture}

\lettrine[nindent=0em,lines=3]{T}{}he suggested edge-oriented backbone architecture is based on X3D framework but with
MobileNet-V3 2D skeleton architecture instead of MobileNet-V2 originally. According to S3D framework the first
convolution has only spatial dimensions ($1 \times 3 \times 3$ kernel) and places the temporal strides in different
positions than the spational one. All the above changes allow us to design the efficient 3D network architecture.

The mentioned 3D architecture constitutes the local path in the described early LGD network design. The global path is
implemented on the same way as in the original paper but with GC block instead of simple GAP operator. The output of the
merged backbone is preserved the same as in X3D+MobileNet-V3 architecture without addition of kernel fusion module.

The backbone output followed the simple spatio-temporal reduction module implemented by GAP operator. On top of network
the Metric-Learning based head is placed. It consists of two consecutive $1 \times 1 \times 1$ convolutions with batch
norms~\cite{BN} and forms the output of the network to be 256-dimensional feature-vector. Note, the network output is
$l_2$ normalized. For more details see the Figure \ref{fig:architecture}.


\section{Network regularization}

\subsection{Multi-head pre-training} \label{pre-training}

\lettrine[nindent=0em,lines=3]{R}{}ecent success of transfer learning approach~\cite{TransferLearning}
through pre-training the network on tremendous general datasets (e.g. ImageNet~\cite{ImageNet} for 2D image
classification and Kinetics~\cite{Kinetics} for the 3D use case) plays important role in the network regularization. It
allows us to train the final model on the limited amount of data by using (aka. transferring) the original pre-trained
weights.

Another point of view on the pre-training stage is formulated as increasing the number of targets~\cite{ImageNet-21k}
to reach the affluent semantics of the learned classes. Unfortunately, the data collection process of video datasets is
time consuming and the existing datasets cannot boast of sufficient number of classes. The solution has been proposed in
the paper~\cite{SplitML} by simultaneously training on the merged set of available datasets. The idea is to sample the
batch from the joint set of samples but split the network heads according to the target datasets. In such framework
each head spawn the independent decision space and there is no possible conflicting between similar classes in
different datasets.

The mentioned above method allows us to use the original annotation of existing datasets but train the network on the
full set of samples simultaneously. The proposed network architecture is trained according to the same idea on the joint
set of Kinetics and YouTube-8M-Segments datasets but with ML-heads (see the next section~\ref{feature-regularzation}).

\subsection{Feature regularization} \label{feature-regularzation}

\lettrine[nindent=0em,lines=3]{U}{}nfortunately, the good network initialization is not enough for training the robust
and accurate final model due to possible over-fitting on the limited-size target dataset. To overcome foregoing issue it
is proposed to use a bag of regularization methods during the training stage.

The paper~\cite{ASLNet} advocate the idea to combine a regular 3D classification network with Metric-Learning (ML)
based design of heads and appropriate ML losses. It adds the two-layer's head on top of a backbone and $l_2$
normalizes the output 256-dimensional feature vector. Additionally, to control the structure of the learned manifold the
extra losses are used: AM-Softmax loss~\cite{AM-Softmax} instead of default CE-loss, CenterPush and LocalPush to model
the sample-sample interactions and the confidence penalty to decrease the impact of overconfident samples. We use the
same approach and for more details see the mentioned above paper.

The ML-based regularization is explicit and it is performed by forcing the extra properties for the decision space. The
different (implicit) way to carry out the regularization is to restrict the expressiveness of features by some rules.
Our experiments made to believe that there are several outstanding methods which are universal enough for the
regularization purposes. The first method is the advanced version of the well-known dropout -- Representation
Self-Challenging (RSC) module~\cite{RSC}. It drops the most relevant filters and thereby forces the network to find the
diverse set of features.

The other regularization method is a part of the AugMix~\cite{AugMix} augmentation. It applies the different
augmentation pipelines to the same sample in the batch and adds auxiliary loss to pull the predictions of the same
instance to each other. Thereby the network tries to learn the augmentation-independent representation. We treat the
method as some kind of self-mutual learning \cite{SFR} and use it as the way to fill the network capacity with utility
task-relevant filters.


\section{Implementation details}

\subsection{Augmentations}

\lettrine[nindent=0em,lines=3]{A}{}s it was described in the previous section the AugMix~\cite{AugMix} augmentation is
used to effectively regularize the network during the final training on the limited-size datasets. Unlike the original
paper we have chosen the fixed set of operations used in the augmentation pipeline: random rotate, crop, horizontal flip
and CrossNorm~\cite{CrossNorm} augmentations. Furthermore, we have found useful to integrate the random selection of
time segment (aka. \textit{temporal crop} during sampling a clip from the full input video) inside the augmentation
pipeline. The last finding significantly increases the difficulty of the auxiliary task.

The other direction of augmenting consists of extending the background diversity of input frames by mixing the source
video clips with some external spatial information. In our experiments the most impressive result have showed the
MixUp augmentation~\cite{MixUp} but adopted for the video input (originally proposed in~\cite{ASLNet}). It selects a
randomly chosen image from the predefined set of images (commonly from ImageNet dataset~\cite{ImageNet}) and mix it
with a full sequence of selected frames in a video clip.

The same model quality can be also achieved by the recently proposed CrossNorm~\cite{CrossNorm} augmentation. To carry
out the last one it is only need at source of pairs of clip mean and variance. Finally, we have chosen the CrossNorm
augmentation in our pipeline due the simplicity for the end user to use that augmentation.

\subsection{Label noise suppression} \label{noise_suppression}

\begin{figure}[ht]
\centering
\includegraphics[width=\textwidth]{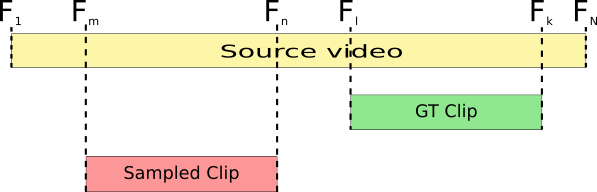}
\caption{Example of invalid clip sampling. $F_1...F_N$ -- source video, $F_l...F_k$ -- ground-truth borders of a target
action clip, $F_m...F_n$ -- incorrectly sampled action clip (according to the common used uniform sampling strategy).}
\label{fig:frame_sampling}
\end{figure}

\lettrine[nindent=0em,lines=3]{O}{}ne more problem for training AR models is related to the \textit{induced label noise}
-- it is possible while a network should be trained on an untrimmed dataset, like ActivityNet~\cite{ActivityNet}. In
this way there is no accurate temporal segmentation of the target action classes and the clip sampling procedure can
select background frames from the video thereby producing the sample with an incorrect label (aka. \textit{induced
label noise} -- see Figure \ref{fig:frame_sampling}). According to the our observations the output label noise can
exceeds more than $90\%$ in a worse case.

There are many solutions have been developed to tackle it, like MARVEL~\cite{MARVEL} and PRISM~\cite{PRISM}.
Unfortunately the mentioned methods reduce the quality in case of relatively clean datasets (e.g. UCF-101
\cite{UCF101}) due to the inability to distinguish the noisy sample from the difficult one. The last thing does not
allow us to use them in the universal framework suitable for all general use cases.

In our opinion, the more elegant way to tackle the mentioned problem is to design the procedure for the careful
clip selection instead of fixing a label of an already sampled clip. In light of this, we have designed the Adaptive
Clip Selection (ACS) procedure. The main idea of the method is to permanently collect the predicted accuracy of each
sampled clip and then sample a new one according to the probability of temporal segment to be a correct carrier of a
target action category.

More formally lets assume that $F=\{i: i \in \overline{1,N}\}$ is set of frame indices in some video and $N$ is a number
of frames in it. The frame sampling procedure aims to select some continues subset of frames with length $n$. We can
split the $F$ into continuous segments $S_1,S_2, ..., S_{N-n+1}$, where $S_i=\{k: i \leq k < i + n \}$. Additionally,
lets associate each frame in a video with some positive score: $V=\{v_i: i \in \overline{1,N}, v_i \geq 0\}$. Having the
probabilities $P_i$ of selecting of each segment $S_i$ we can sample it from the defined above multinomial distribution.
In this way $P_i$ can be expressed as follow:

\begin{equation}
\label{eq:segment_prob}
P_i=\frac{\sum_{k \in S_i} v_kI[v_k \geq 0] + 1}{\sum_t \sum_{k \in S_t} v_kI[v_k \geq 0] + 1}
\end{equation}

During the training procedure we store the predicted scores of classes after each forward pass. Lets assume the segment
$S_i$ has been predicted as class $\overline{y_i}$ with the probability $\overline{P_i}$. The true label of segment
$S_i$ is $y_i$. To enable the collection of statistics we update the associated with frames scores according to the
following equation:

\begin{equation}
\label{eq:score_update}
v_{k}^{new} = v_{k}^{old} + (I[\overline{y_i} = y_i] - I[\overline{y_i} \neq y_i])P_i, k \in S_i
\end{equation}

In practice we enable ACS for each video sample in dataset independently and do it after collecting the sufficient
statistics of frames -- number of frames with non-zero score should be not less than some threshold (70\% in our
experiments). Before that the ACS is disabled and clip sampling is performed uniformly. Note that collecting statistics
is performed from the beginning to the end of training (except the very first epoch to omit the potential volatility of
the beginning). Also we have experimented with labeling the sample as ignore (or negative) in case of all frames in a
video predicted as incorrect (all $v_i < 0$) but the quality was worse because the method faced with the same as
previous label correction methods problem -- the mentioned above dilemma of a sample difficulty or a label noise.

\subsection{Training}

\lettrine[nindent=0em,lines=3]{T}{}he network training procedure is performed in multi-stage manner to preserve the
pre-trained weights (see section~\ref{pre-training}) and speed up training at all. Training stages are as follow:

\begin{enumerate}
\item Training the network head only. During this stage we freeze the whole backbone parameters (all batch norms in
backbone are switched to the inference mode) and train the head only with high learning rate ($10^{-2}$). That training
allows to roughly optimize the class centroids and prevents the future gradient explosion after enabling the gradient
propagation for all model parameters.
\item On the next stage we enable learning rate WarmUp~\cite{WarmUp}. It increases the learning rate from $10^{-5}$ up
to $10^{-3}$ by cosine schedule. It allows to sew together the gradients of the head and the rest network.
\item Finally the training procedure is switched to the default one.
\end{enumerate}

The main training procedure is performed by cosing learning rate schedule starting from $10^{-3}$ and ending with
$10^{-5}$. We use slightly modified version of cosine schedule to increase a fraction of the time on a high learning
rate. Practically it prevents from strong over-fitting on the late phase of training in case of limited-size datasets.
The learning rate at the $t$-th iteration is given by the following expression:

\begin{equation}
\label{eq:lr_schedule}
\eta_t=\eta_{min} + \frac{1}{2}(\eta_{max} - \eta_{min})(1 + cos\left(\left(\frac{T_{cur}}{T_{max}}\right)^{\alpha} \pi \right)
\end{equation}

\noindent where $\eta_{min}$ and $\eta_{max}$ are the lower and upper bounds respectively for the learning rate.
$T_{cur}$ represents the current iteration and $T_{max}$ is maximum number of iterations. The parameter $\alpha$
allows us to control the fraction of high learning rate phase -- in our experiments we have fixed it to the $1.5$ value.

Furthermore, it was found that training with the recently proposed Adaptive Gradient Clipping (AGC)~\cite{AGC} allows us
to not only speed up training but increase the model accuracy too. In all experiments below we set it enabled for our
network. The final network is trained on two GPUs by 12 clips per node with SGD optimizer and weight decay
regularization using the PyTorch framework~\cite{PyTorch}.


\begin{table*}[t]
\caption{The pivot table of different frame sampling strategies and test protocols. For each combination the Top-1
accuracy is reported.}
\label{table:ablation_samplers}
\centering
\begin{tabular}{l|cc|cc|cc}
\multirow{2}{*}{\textbf{Sampler}}& \multicolumn{2}{c|}{\textbf{UCF-101}}     & \multicolumn{2}{c|}{\textbf{ActivityNet-200}} & \multicolumn{2}{c}{\textbf{Jester-27}}    \\
                                 & \textbf{1 segment} & \textbf{10 segments} & \textbf{1 segment}   & \textbf{10 segments}   & \textbf{1 segment} & \textbf{10 segments} \\ \hline
sparse                           & 93.79              & 94.71                & 59.94                & 74.32                  & 95.25              & 95.68                \\
continuous                       & 93.63              & 94.85                & 64.19                & 75.27                  & 95.43              & 95.56
\end{tabular}
\end{table*}

\section{Experimental Results}

\subsection{Data}

\lettrine[nindent=0em,lines=3]{W}{}e conduct experiments on several commonly-used benchmarks for general action and
gesture recognition. The listed below datasets allow us to validate the proposed architecture in different scenarios:
trimmed and untrimmed general purpose action recognition and continuous hand gesture recognition. As it was described
early the first category checks the ability to learn the appearance-based action recognition while the last one -- the
ability to model the motion component of actions.

\begin{itemize}
\item \textbf{UCF-101}~\cite{UCF101} is a middle-size general-purpose action recognition dataset of trimmed action
videos, splitted on the 101 action categories. The total number of samples is 13320 and there are three train-val splits
provided. As many other researchers we report the results on the first split only to reduce the training time.
\item \textbf{ActivityNet-200}~\cite{ActivityNet} is a middle-size general-purpose action recognition dataset. Unlike
the previous dataset the source videos are untrimmed and split on 200 action categories. Note that the dataset is
designed to solve the temporal action detection/segmentation task mostly but we adopt it for action recognition purposes 
(see the next section for more details). The total number of available for downloading video instances is 17196.
\item \textbf{Jester-27}~\cite{Jester} is large-size dataset of labeled video clips that shows humans performing
pre-defined hand gestures. The dataset consists of 148092 unique video clips splitted on the 27 gesture categories.
\end{itemize}

\subsection{Evaluation Protocol}

\lettrine[nindent=0em,lines=3]{T}{}he main protocol to measure the performance of action recognition algorithms consists
of top-1 and top-5 accuracies. The last metric is used to flatten the perturbations in case of noisy labels in the
annotation. However as it was mentioned in the paper~\cite{ASLNet} the transition to ML-based heads allows us to measure
more noise resistant metric, like rank mAP (for more details see the reported above paper). In this paper we measure it
too.

Another question is related to the choice of a procedure for the clip selection from an input video during the testing
phase. Commonly used approach is to split the input video onto 10 temporal segments and apply a AR network for each
segment independently. Sometimes AR network is run for several crops inside each temporal segment. For the end user it
means that the reported performance metrics in terms of GFlops should be multiplied onto 30 (10 segments $\times$ 3
crops). The mentioned method is suitable for the academic community (to show the best quality) but not for a business.
To report the fair performance-accuracy pairs we follow the reduced protocol -- single central spatio-temporal crop for
each input video (for comparison purposes we report the metrics for both protocols in the section~\ref{ablation_study}).
Note, the ActivityNet data is untrimmed, so the single temporal crop may not reflect the target class due to temporal
mismatch with ground-truth (see the section~\ref{noise_suppression}). So, the 10-temporal-crop protocol is considered as
a primary one for the ActivityNet benchmark.

One more possible difference in measurements is related to the frame sampling procedure. For the general-purpose
datasets it is reasonable to sample frames inside a segment uniformly to achieve the best temporal coverage.
Unfortantely, the discussed method is not suitable for the real-time applications (e.g. hand gesture recognition) which
process frames on-the-fly and does not have an access to the future frames. In the paper~\cite{ASLNet} the continues
protocol has been proposed -- sampling the frames during the train and test phases with a fixed frame-rate. In the paper
we follow the same continues protocol (the comparison between protocols can be found in the~\ref{ablation_study}
section).

\begin{table}[h]
\caption{The ablation study of the backbone initialization on the UCF-101 dataset.}
\label{table:ablation_init}
\centering
\begin{tabular}{l|ccc}
\textbf{Method}           & \textbf{Top-1} & \textbf{Top-5} & \textbf{mAP} \\ \hline
from scratch              & 64.05          & 87.97          & 57.33        \\
ImageNet (2D init)        & 71.90          & 90.67          & 77.74        \\
Kinetics (3D init)        & 92.81          & 98.76          & 96.88        \\
Kinetics+Ytb8M (3D init)  & 93.63          & 99.10          & 97.82   
\end{tabular}
\end{table}

\subsection{Ablation study} \label{ablation_study}

\lettrine[nindent=0em,lines=3]{B}{}elow we present the ablation study of the proposed framework in terms of the frame
sampling strategy (comparison between sparse and continues protocols during the inference), the pre-training method and
the influence of the introduced ACS module.

\begin{table*}[h]
\caption{The comparison of using ACS method on UCF-101 and ActivityNet-200 datasets. Note the metrics on the ActivityNet
         dataset is specified for the testing protocol with 10 temporal segments.}
\label{table:ablation_acs}
\centering
\begin{tabular}{l|ccc|ccc}
\multicolumn{1}{c|}{\multirow{2}{*}{\textbf{Method}}} & \multicolumn{3}{c|}{\textbf{UCF-101}}               & \multicolumn{3}{c}{\textbf{ActivityNet-200}} \\
\multicolumn{1}{c|}{}                                 & \textbf{Top-1} & \textbf{Top-5} & \textbf{mAP} & \textbf{Top-1} & \textbf{Top-5} & \textbf{mAP} \\ \hline
w/o ACS                                               & 93.52          & 99.07          & 97.50        & 74.89          & 92.77          & 73.33        \\
w/ ACS                                                & 93.63          & 99.10          & 97.82        & 75.27          & 92.70          & 73.81            
\end{tabular}
\end{table*}

\textbf{Frame sampling strategy.} First, we would like to compare the frame sampling strategies. Depending on the target
use case there are two possible solutions described early: sparse sampling and fixed frame rate sampling (or just with
fixed temporal stride). The first strategy assumes the low impact of an ratio between a video and the network input
lengths but is not suitable for the live demo mode due to inability to see the future frames. Recently the sparse
sampling method~\cite{MoViNet} allowed the authors to show the magnificent performance with single temporal crop. Our
experiments (see the Table~\ref{table:ablation_samplers}) demonstrate that two-path network with separated global
context branch is able to reduce the impact of an sampling strategy. Moreover we see the improvement in case of single
crop for the ActivityNet dataset. In our opinion it is because the target clip with a valid action is significantly
shorter than the full video and as a result the sparse sampling strategy collects a scarce number of valid frames.
Generally speaking, the LIGAR framework allows us to close the question of the frame sampling strategy in case of live demo
applications.

Additionally, we have compared here the difference between testing protocols, especially the single- or multi-view video
prediction. As it is expected increasing the number of views per an video improves the accuracy metric due to smoothing
the impact of possible temporal mismatch between the unknown ground truth and the tested central temporal crop.
Unfortunately most of papers report the multi-view metrics to get the best results and it is not suitable for the live
demo scenario. Regarding the reported results the difference between the two testing protocols is comfortable with the
exception of ActivityNet dataset (the reason of that is described early).

\begin{table*}[ht]
\caption{Comparison the proposed LIGAR framework with SOTA solutions on UCF-101 and Jester-27 datasets. For fairness we
report the number of views for each measure if it is specified.}
\label{table:results}
\centering
\begin{tabular}{l|c|c|c|c|ccc|ccc}
\multicolumn{1}{c|}{\multirow{2}{*}{\textbf{Name}}} & \multirow{2}{*}{\textbf{\begin{tabular}[c]{@{}c@{}}Input\\ frames\end{tabular}}} & \multirow{2}{*}{\textbf{Views}} & \multirow{2}{*}{\textbf{\begin{tabular}[c]{@{}c@{}}Single\\ GFLOPs\end{tabular}}} & \multirow{2}{*}{\textbf{MParams}} & \multicolumn{3}{c|}{\textbf{UCF-101}} & \multicolumn{3}{c}{\textbf{Jester-27}} \\
\multicolumn{1}{c|}{} &  &  &  &  & \textbf{Top-1} & \textbf{Top-5} & \textbf{mAP} & \textbf{Top1} & \textbf{Top-5} & \textbf{mAP} \\ \hline
R(2+1)D-BERT~\cite{Bert} & 64 & - & 152.97 & 66.67 & 98.69 & - & - & - & - & - \\
LGD-3D RGB~\cite{LGD} & 16 & 15 & - & - & 97.00 & - & - & - & - & - \\
PAN ResNet101~\cite{PAN} & 32 & 2 & 251.7 & - & - & - & - & 97.40 & 99.90 & - \\
STM~\cite{STM} & 16 & 10 & 66.5 & 22.4 & 96.20 & - & - & 96.70 & 99.90 & - \\
3D-MobileNetV2 1.0x~\cite{3DMob} & 16 & 10 & 0.45 & 3.12 & 81.60 & - & - & 94.59 & - & - \\ \hline
\multirow{2}{*}{LIGAR (our)} & \multirow{2}{*}{16} & 1 & \multirow{2}{*}{4.74} & \multirow{2}{*}{4.47} & 93.63 & 99.10 & 97.82 & 95.43 & 99.45 & 96.95 \\
 &  & 10 &  &  & 94.85 & 99.50 & 98.61 & 95.56 & 99.52 & 97.33
\end{tabular}
\end{table*}

\textbf{Network initialization.} Another important question is about the network initialization and noteworthiness of a
pre-training stage. Originally, pre-training is designed to transfer the knowledge from the big datasets to the small
one thereby reducing a harmful effect of over-fitting. For the 3D-based networks the initialization have two sources: 2D
initialization of a backbone only before the inflating procedure (see the I3D~\cite{I3D} paper for more details) and
direct 3D initialization of a full network. In the Table~\ref{table:ablation_init} we have summarized the potential
strategies of the network initialization. Note, the last line is different from the previous one by enabling the
mentioned early multi-head pre-training on the merged dataset. Overall, the model behavior reflects the intuition
behind it -- the method with more data during the pre-training stage exceeds the previous one in terms of all measured
metrics.

\textbf{Induced noise suppression.} The last question is related to the suppression of the described early induced label
noise. The main benchmark to measure the importance of the proposed Adaptive Clip Selection (ACS) module is ActivityNet
dataset. As it was mentioned before we do not follow the original ActivityNet protocol and measure the action
recognition metrics instead of localization one. Unfortunately we cannot measure the real performance of the model on
this dataset due to the lack of an accurate temporal annotation of target actions and it is expected that the real
quality is higher than it is announced. Nevertheless, we can see in the Table~\ref{table:ablation_acs} the improvement
over the baseline by using the proposed ACS module. Furthermore, the improvement is observed for the initially clean
dataset like UCF-101. In our opinion, it is because the impact of induced label noise is much stronger than it is
commonly believed even for the clean popular datasets.

\subsection{Comparison with the State-of-the-Arts}

\lettrine[nindent=0em,lines=3]{W}{}e further demonstrate the advances of our proposed LIGAR framework in comparison with
state-of-the-art methods for the general-purpose action recognition. For fair comparison all methods use the RGB
modality as an input. We report both results for possible testing protocols -- 1 and 10 temporal crops (spatial crops do
not benefit in our experiments). In the Table~\ref{table:results} we have collected the best solutions in term of
accuracy on two sufficiently different datasets.

\textbf{Results on UCF-101.} We first verify the appearance modeling ability on UCF-101. The model can achieve the
superior results compared with other methods which are computationally more expensive. For example the BERT-like
solution~\cite{Bert} is $\times 32$ times more expensive in term of GFLOPs but the accuracy drop is only $5\%$ of top-1
metric. In case of equal testing protocol (10 views per video) the drop is slightly less. In comparison to the solution
with a similar computation budget~\cite{3DMob} the accuracy of the reported approach is significantly better.

\textbf{Results on Jester-27.} We also compare with other methods on Jester-27 to verify the model ability to make the
prediction according to the motion component. Comparing with results on general dataset the gap between heavy solutions
and the proposed one is even less. Specifically, we can observe that our proposed method drops less than 2 percentage in
comparison to the SOTA solution~\cite{PAN}. Like for the previous dataset the advantage in terms of a computation budget
even more than 53 times.


\section{Conclusion}

\lettrine[nindent=0em,lines=3]{T}{}his paper has presented the extension to the LGD framework for more efficient and
accurate solving an action recognition problem for a wide range of applications, like general (appearance-based) and
gesture (motion-based) recognition problems. Moreover, the paper proposed a novel clips selection module to tackle
the induced label noise issues. The described training pipeline allows us to train a robust DL-based solution which is able
to solve most of real-world action recognition problems in a fast and accurate manner. The reported results give us the
hope that DL-based solutions will continue advance on the rest vital challenges of a humanity.


\bibliographystyle{splncs}
\bibliography{egbib}

\begin{thebibliography}{10}

\bibitem{MN-V1}
Howard, A.G., Zhu, M., Chen, B., Kalenichenko, D., Wang, W., Weyand, T.,
  Andreetto, M., Adam, H.:
\newblock Mobilenets: Efficient convolutional neural networks for mobile vision
  applications.
\newblock CoRR \textbf{abs/1704.04861} (2017)

\bibitem{MN-V2}
Sandler, M., Howard, A.G., Zhu, M., Zhmoginov, A., Chen, L.:
\newblock Inverted residuals and linear bottlenecks: Mobile networks for
  classification, detection and segmentation.
\newblock CoRR \textbf{abs/1801.04381} (2018)

\bibitem{MN-V3}
Howard, A., Sandler, M., Chu, G., Chen, L., Chen, B., Tan, M., Wang, W., Zhu,
  Y., Pang, R., Vasudevan, V., Le, Q.V., Adam, H.:
\newblock Searching for mobilenetv3.
\newblock CoRR \textbf{abs/1905.02244} (2019)

\bibitem{OSNet}
Zhou, K., Yang, Y., Cavallaro, A., Xiang, T.:
\newblock Learning generalisable omni-scale representations for person
  re-identification.
\newblock CoRR \textbf{abs/1910.06827} (2019)

\bibitem{I3D}
Carreira, J., Zisserman, A.:
\newblock Quo vadis, action recognition? {A} new model and the kinetics
  dataset.
\newblock CoRR \textbf{abs/1705.07750} (2017)

\bibitem{P3D}
Qiu, Z., Yao, T., Mei, T.:
\newblock Learning spatio-temporal representation with pseudo-3d residual
  networks.
\newblock CoRR \textbf{abs/1711.10305} (2017)

\bibitem{R(2+1)D}
Tran, D., Wang, H., Torresani, L., Ray, J., LeCun, Y., Paluri, M.:
\newblock A closer look at spatiotemporal convolutions for action recognition.
\newblock CoRR \textbf{abs/1711.11248} (2017)

\bibitem{S3D}
Xie, S., Sun, C., Huang, J., Tu, Z., Murphy, K.:
\newblock Rethinking spatiotemporal feature learning for video understanding.
\newblock CoRR \textbf{abs/1712.04851} (2017)

\bibitem{X3D}
Feichtenhofer, C.:
\newblock {X3D:} expanding architectures for efficient video recognition.
\newblock CoRR \textbf{abs/2004.04730} (2020)

\bibitem{Kinetics}
Carreira, J., Noland, E., Hillier, C., Zisserman, A.:
\newblock A short note on the kinetics-700 human action dataset.
\newblock CoRR \textbf{abs/1907.06987} (2019)

\bibitem{Ytb8m}
Abu{-}El{-}Haija, S., Kothari, N., Lee, J., Natsev, P., Toderici, G.,
  Varadarajan, B., Vijayanarasimhan, S.:
\newblock Youtube-8m: {A} large-scale video classification benchmark.
\newblock CoRR \textbf{abs/1609.08675} (2016)

\bibitem{TSN}
Wang, L., Xiong, Y., Wang, Z., Qiao, Y., Lin, D., Tang, X., Gool, L.V.:
\newblock Temporal segment networks: Towards good practices for deep action
  recognition.
\newblock CoRR \textbf{abs/1608.00859} (2016)

\bibitem{TSM}
Lin, J., Gan, C., Han, S.:
\newblock Temporal shift module for efficient video understanding.
\newblock CoRR \textbf{abs/1811.08383} (2018)

\bibitem{TRN}
Zhou, B., Andonian, A., Torralba, A.:
\newblock Temporal relational reasoning in videos.
\newblock CoRR \textbf{abs/1711.08496} (2017)

\bibitem{LGD}
Qiu, Z., Yao, T., Ngo, C., Tian, X., Mei, T.:
\newblock Learning spatio-temporal representation with local and global
  diffusion.
\newblock CoRR \textbf{abs/1906.05571} (2019)

\bibitem{C3D}
Tran, D., Bourdev, L.D., Fergus, R., Torresani, L., Paluri, M.:
\newblock {C3D:} generic features for video analysis.
\newblock CoRR \textbf{abs/1412.0767} (2014)

\bibitem{TwoStream}
Feichtenhofer, C., Pinz, A., Zisserman, A.:
\newblock Convolutional two-stream network fusion for video action recognition.
\newblock CoRR \textbf{abs/1604.06573} (2016)

\bibitem{TRG}
Tran, D., Bourdev, L.D., Fergus, R., Torresani, L., Paluri, M.:
\newblock {C3D:} generic features for video analysis.
\newblock CoRR \textbf{abs/1412.0767} (2014)

\bibitem{SlowFast}
Feichtenhofer, C., Fan, H., Malik, J., He, K.:
\newblock Slowfast networks for video recognition.
\newblock CoRR \textbf{abs/1812.03982} (2018)

\bibitem{AssembleNet}
Ryoo, M.S., Piergiovanni, A.J., Tan, M., Angelova, A.:
\newblock Assemblenet: Searching for multi-stream neural connectivity in video
  architectures.
\newblock CoRR \textbf{abs/1905.13209} (2019)

\bibitem{RCCA}
Cao, C., Lu, Y., Zhang, Y., Jiang, D., Zhang, Y.:
\newblock Efficient spatialtemporal context modeling for action recognition.
\newblock CoRR \textbf{abs/2103.11190} (2021)

\bibitem{NL}
Wang, X., Girshick, R.B., Gupta, A., He, K.:
\newblock Non-local neural networks.
\newblock CoRR \textbf{abs/1711.07971} (2017)

\bibitem{Bert}
Kalfaoglu, M.E., Kalkan, S., Alatan, A.A.:
\newblock Late temporal modeling in 3d {CNN} architectures with {BERT} for
  action recognition.
\newblock CoRR \textbf{abs/2008.01232} (2020)

\bibitem{ShuffleNet-V1}
Zhang, X., Zhou, X., Lin, M., Sun, J.:
\newblock Shufflenet: An extremely efficient convolutional neural network for
  mobile devices.
\newblock CoRR \textbf{abs/1707.01083} (2017)

\bibitem{ShuffleNet-V2}
Ma, N., Zhang, X., Zheng, H., Sun, J.:
\newblock Shufflenet {V2:} practical guidelines for efficient {CNN}
  architecture design.
\newblock CoRR \textbf{abs/1807.11164} (2018)

\bibitem{EfficientNet}
Tan, M., Le, Q.V.:
\newblock Efficientnet: Rethinking model scaling for convolutional neural
  networks.
\newblock CoRR \textbf{abs/1905.11946} (2019)

\bibitem{MoViNet}
Kondratyuk, D., Yuan, L., Li, Y., Zhang, L., Tan, M., Brown, M., Gong, B.:
\newblock Movinets: Mobile video networks for efficient video recognition.
\newblock CoRR \textbf{abs/2103.11511} (2021)

\bibitem{MixUp}
Zhang, H., Ciss{\'{e}}, M., Dauphin, Y.N., Lopez{-}Paz, D.:
\newblock mixup: Beyond empirical risk minimization.
\newblock CoRR \textbf{abs/1710.09412} (2017)

\bibitem{AugMix}
Hendrycks, D., Mu, N., Cubuk, E.D., Zoph, B., Gilmer, J., Lakshminarayanan, B.:
\newblock Augmix: A simple data processing method to improve robustness and
  uncertainty (2020)

\bibitem{CrossNorm}
Tang, Z., Gao, Y., Zhu, Y., Zhang, Z., Li, M., Metaxas, D.N.:
\newblock Selfnorm and crossnorm for out-of-distribution robustness.
\newblock CoRR \textbf{abs/2102.02811} (2021)

\bibitem{Dropout}
Srivastava, N., Hinton, G., Krizhevsky, A., Sutskever, I., Salakhutdinov, R.:
\newblock Dropout: A simple way to prevent neural networks from overfitting.
\newblock Journal of Machine Learning Research \textbf{15}(56) (2014)
  1929--1958

\bibitem{ContinuousDropout}
Shen, X., Tian, X., Liu, T., Xu, F., Tao, D.:
\newblock Continuous dropout.
\newblock CoRR \textbf{abs/1911.12675} (2019)

\bibitem{InfoDrop}
Shi, B., Zhang, D., Dai, Q., Zhu, Z., Mu, Y., Wang, J.:
\newblock Informative dropout for robust representation learning: {A}
  shape-bias perspective.
\newblock CoRR \textbf{abs/2008.04254} (2020)

\bibitem{FocusedDropout}
Xie, T., Liu, M., Deng, J., Cheng, X., Wang, X., Liu, M.:
\newblock Focuseddropout for convolutional neural network.
\newblock CoRR \textbf{abs/2103.15425} (2021)

\bibitem{MutualLearning}
Zhang, Y., Xiang, T., Hospedales, T.M., Lu, H.:
\newblock Deep mutual learning.
\newblock CoRR \textbf{abs/1706.00384} (2017)

\bibitem{SFR}
Hou, Z., Fan, W.:
\newblock A new training framework for deep neural network.
\newblock CoRR \textbf{abs/2103.07350} (2021)

\bibitem{ASLNet}
Izutov, E.:
\newblock {ASL} recognition with metric-learning based lightweight network.
\newblock CoRR \textbf{abs/2004.05054} (2020)

\bibitem{AM-Softmax}
Wang, F., Liu, W., Liu, H., Cheng, J.:
\newblock Additive margin softmax for face verification.
\newblock CoRR \textbf{abs/1801.05599} (2018)

\bibitem{RMNet}
Izutov, E.:
\newblock Fast and accurate person re-identification with rmnet.
\newblock CoRR \textbf{abs/1812.02465} (2018)

\bibitem{MSASL}
Joze, H.R.V., Koller, O.:
\newblock {MS-ASL:} {A} large-scale data set and benchmark for understanding
  american sign language.
\newblock CoRR \textbf{abs/1812.01053} (2018)

\bibitem{UCF101}
Soomro, K., Zamir, A.R., Shah, M.:
\newblock {UCF101:} {A} dataset of 101 human actions classes from videos in the
  wild.
\newblock CoRR \textbf{abs/1212.0402} (2012)

\bibitem{ActivityNet}
Heilbron, F.C., Escorcia, V., Ghanem, B., Niebles, J.C.:
\newblock Activitynet: A large-scale video benchmark for human activity
  understanding.
\newblock In: 2015 IEEE Conference on Computer Vision and Pattern Recognition
  (CVPR). (2015)  961--970

\bibitem{GCN}
Cao, Y., Xu, J., Lin, S., Wei, F., Hu, H.:
\newblock Global context networks.
\newblock CoRR \textbf{abs/2012.13375} (2020)

\bibitem{BN}
Ioffe, S., Szegedy, C.:
\newblock Batch normalization: Accelerating deep network training by reducing
  internal covariate shift.
\newblock CoRR \textbf{abs/1502.03167} (2015)

\bibitem{TransferLearning}
Zhuang, F., Qi, Z., Duan, K., Xi, D., Zhu, Y., Zhu, H., Xiong, H., He, Q.:
\newblock A comprehensive survey on transfer learning.
\newblock CoRR \textbf{abs/1911.02685} (2019)

\bibitem{ImageNet}
Russakovsky, O., Deng, J., Su, H., Krause, J., Satheesh, S., Ma, S., Huang, Z.,
  Karpathy, A., Khosla, A., Bernstein, M., Berg, A.C., Fei-Fei, L.:
\newblock Imagenet large scale visual recognition challenge.
\newblock International Journal of Computer Vision (IJCV) \textbf{115}(3)
  (2015)  211--252

\bibitem{ImageNet-21k}
Ridnik, T., Baruch, E.B., Noy, A., Zelnik{-}Manor, L.:
\newblock Imagenet-21k pretraining for the masses.
\newblock CoRR \textbf{abs/2104.10972} (2021)

\bibitem{SplitML}
Sanakoyeu, A., Tschernezki, V., B{\"{u}}chler, U., Ommer, B.:
\newblock Divide and conquer the embedding space for metric learning.
\newblock CoRR \textbf{abs/1906.05990} (2019)

\bibitem{RSC}
Huang, Z., Wang, H., Xing, E.P., Huang, D.:
\newblock Self-challenging improves cross-domain generalization.
\newblock CoRR \textbf{abs/2007.02454} (2020)

\bibitem{MARVEL}
Lin, J.Z., Bradic, J.:
\newblock Learning to combat noisy labels via classification margins.
\newblock CoRR \textbf{abs/2102.00751} (2021)

\bibitem{PRISM}
Liu, C., Yu, H., Li, B., Shen, Z., Gao, Z., Ren, P., Xie, X., Cui, L., Miao,
  C.:
\newblock Noise-resistant deep metric learning with ranking-based instance
  selection.
\newblock CoRR \textbf{abs/2103.16047} (2021)

\bibitem{WarmUp}
Goyal, P., Doll{\'{a}}r, P., Girshick, R.B., Noordhuis, P., Wesolowski, L.,
  Kyrola, A., Tulloch, A., Jia, Y., He, K.:
\newblock Accurate, large minibatch {SGD:} training imagenet in 1 hour.
\newblock CoRR \textbf{abs/1706.02677} (2017)

\bibitem{AGC}
Brock, A., De, S., Smith, S.L., Simonyan, K.:
\newblock High-performance large-scale image recognition without normalization.
\newblock CoRR \textbf{abs/2102.06171} (2021)

\bibitem{PyTorch}
Paszke, A., Gross, S., Massa, F., Lerer, A., Bradbury, J., Chanan, G., Killeen,
  T., Lin, Z., Gimelshein, N., Antiga, L.,  et~al.:
\newblock Pytorch: An imperative style, high-performance deep learning library.
\newblock In: Advances in Neural Information Processing Systems. (2019)
  8024--8035

\bibitem{Jester}
Materzynska, J., Berger, G., Bax, I., Memisevic, R.:
\newblock The jester dataset: A large-scale video dataset of human gestures.
\newblock In: 2019 IEEE/CVF International Conference on Computer Vision
  Workshop (ICCVW). (2019)  2874--2882

\bibitem{PAN}
Zhang, C., Zou, Y., Chen, G., Gan, L.:
\newblock {PAN:} towards fast action recognition via learning persistence of
  appearance.
\newblock CoRR \textbf{abs/2008.03462} (2020)

\bibitem{STM}
Jiang, B., Wang, M., Gan, W., Wu, W., Yan, J.:
\newblock {STM:} spatiotemporal and motion encoding for action recognition.
\newblock CoRR \textbf{abs/1908.02486} (2019)

\bibitem{3DMob}
K{\"{o}}p{\"{u}}kl{\"{u}}, O., Kose, N., Gunduz, A., Rigoll, G.:
\newblock Resource efficient 3d convolutional neural networks.
\newblock CoRR \textbf{abs/1904.02422} (2019)

\end{thebibliography}


\end{document}